%% file: preprint.tex
\pdfoutput=1
\RequirePackage[svgnames,table]{xcolor}

\documentclass[11pt,letterpaper]{template/style}

\input{template/preamble}
\input{template/command}

\usepackage[capitalize]{cleveref}
\usepackage{flafter} 

\definecolor{cvprblue}{rgb}{0.21,0.49,0.74}
\definecolor{duoblue}{RGB}{46,110,165}
\definecolor{duoorange}{RGB}{196,112,58}

\title{Duo: Towards Spatiotemporal Action Coordination in Bimanual Policies}

\author[1,*]{Zhimu Zhou}
\author[2,*]{Yuqi Yang}
\author[1]{Yixin Lin}
\author[3]{Junsen Zhang}
\author[4]{Jian Zhao}
\author[5]{Jiayu Zhong}
\author[6]{Jiaxun Sun}
\author[7]{Bingxin Wang}
\author[2,\dag]{Binhua Zuo}
\author[1,\dag]{Bo Zhao}
\affil[1]{Shanghai Jiao Tong University}
\affil[2]{Quicrobot}
\affil[3]{Zhejiang University}
\affil[4]{Tsinghua University}
\affil[5]{Technical University of Munich}
\affil[6]{ETH Zurich}
\affil[7]{The Hong Kong University of Science and Technology\vspace{10pt}}

\providecommand{\maketitlesupplementary}{\clearpage\appendix\section*{Appendix}}

\newcommand{\DuoPreprintMode}{}

\fancypagestyle{firststyle}{%
  \fancyhead[L]{\small Preprint. Under review.}%
  \fancyhead[R]{%
    \ifdefined\paperurl
      \if\relax\the\paperurl\relax\else
        \href{\the\paperurl}{\urlheaderfont \itshape \the\paperurl}\\%
      \fi
    \fi
  }%
  
  \fancyfoot[L]{\footerfont\textsuperscript{*}Equal contribution.\quad\textsuperscript{\dag}Corresponding author.}%
  \fancyfoot[R]{\footerfont\bfseries\relax}%
  \fancyfoot[C]{\footerfont\bfseries\relax}%
}
\begin{document}
\input{sec/0_abstract}
\maketitle

\input{sec/1_intro}
\input{sec/2_related_work}
\input{sec/3_method}
\input{sec/4_evaluation}

\input{sec/5_results}

\input{sec/6_ablations}
\input{sec/7_conclusion}
\input{sec/8_future}

{%
\small
\bibliographystyle{abbrvnat}
\bibliography{main}
}

\input{sec/X_suppl}

\end{document}

%% file: template/preamble.tex
\usepackage{graphicx}
\usepackage{float,epstopdf}
\usepackage{bbm}

\usepackage{microtype}

\usepackage[numbers]{natbib}
\setcitestyle{square}

\usepackage{subcaption}
\usepackage{booktabs}

\usepackage{amsmath}
\usepackage{amssymb}
\usepackage{mathtools}
\usepackage{amsthm}
\usepackage{dsfont}
\usepackage{makecell}
\usepackage{multirow}
\usepackage{amsfonts}
\usepackage{mathrsfs}
\usepackage[amssymb, thickqspace]{SIunits}
\DeclareFontFamily{U}{rsfs}{\skewchar\font127}
\DeclareFontShape{U}{rsfs}{m}{n}{%
  <5> <5.5> <6> rsfs5
  <7> rsfs7
  <8> <9> <10> <10.95> <12> <14.4> <17.28> <20.74> <24.88> rsfs10
}{}
\DeclareFontShape{OML}{eur}{m}{n}{%
  <5> <5.5> <6> <7> <8> <9> gen * eurm
  <10> <10.95> <12> <14.4> <17.28> <20.74> <24.88> eurm10
}{}
\usepackage{enumitem}
\usepackage{pgfplotstable}
\pgfplotsset{compat=1.18}
\usepackage{lipsum}		

\usepackage{microtype}
\usepackage{graphicx}
\usepackage{booktabs} 
\usepackage[table]{xcolor}
\usepackage[normalem]{ulem} 

\usepackage{cases}
\usepackage{wrapfig}

\usepackage{url}

\definecolor{huskypurple}{HTML}{4B2E83}

\usepackage{titletoc}

\usepackage{listings}
\lstdefinestyle{promptstyle}{
  basicstyle=\ttfamily\footnotesize,
  breaklines=true,
  breakautoindent=false,
  breakindent=0pt,
  postbreak=\mbox{\textcolor{gray}{$\hookrightarrow$}\space},
  columns=fullflexible,
  keepspaces=true,
  frame=single,
  framesep=5pt,
  xleftmargin=6pt,
  xrightmargin=6pt,
  aboveskip=8pt,
  belowskip=8pt,
  showstringspaces=false,
}

\usepackage{fontawesome5}   

%% file: template/command.tex
\makeatletter
\def\munderbar#1{\underline{\sbox\tw@{$#1$}\dp\tw@\z@\box\tw@}}
\makeatother

\AddToHook{cmd/appendix/before}{%
  \setcounter{axiom}{0}%
}

\newcommand{\be}{\begin{equation}}
\newcommand{\ee}{\end{equation}}
\newcommand{\bea}{\begin{equation*}\begin{aligned}}
\newcommand{\eea}{\end{aligned}\end{equation*}}



%% file: sec/0_abstract.tex
\begin{abstract}
Bimanual manipulation requires each arm to perform distinct motions while coordinating with the other.
Common flow-based robotic policies combine both arms' actions into one token per future timestep, giving attention distinct positions across time but not across arms.
We introduce \textbf{Duo}, a simple but effective design that models spatial and temporal dependencies within one shared action head. 
Specifically, Duo separates both arms into two action tokens per timestep (\Cref{fig:duovla_comparison}), allowing each arm's motion to interact via cross-arm attention during flow matching. 
On RoboTwin 2.0, Duo improves the overall success of all four base policies, from 0.45B to 3.62B parameters, with gains of 12.55 and 9.02 percentage points for FLOWER (0.89B) and $\pi_{0.5}$ (3.62B), respectively.
Despite its compact size, FLOWER-Duo achieves the best randomized-generalization, precision, and memory scores on RoboDojo. 
In real-world tasks, it outperforms $\pi_{0.5}$ in both success rate and execution speed, while scaling effectively with increasing training data.
To provide deeper insight into the design of Duo, we conduct systematic ablations on inter-arm coordination, action grouping, and token granularity, confirming that cross-arm attention drives the improvement while arm-wise interaction offers the best accuracy–compute trade-off.

\end{abstract}

%% file: sec/1_intro.tex
\section{Introduction}
\label{sec:intro}

\input{figure/0_comparison}

Bimanual manipulation enables robots to handle tasks that a single arm cannot finish independently~\cite{zhao2023act,grotz2024peract2}, such as flipping a parcel, handling a deformable mailer, or folding a cardboard box.
These tasks are challenging because they require spatial alignment and temporal coordination between the arms~\cite{krebs2022taxonomy,liu2024rdt,lv2025kstar}.
For example, in a handover, one arm slides a parcel to the center and flips it, while the other arm must position itself at the parcel's updated pose (spatial alignment) and sweep it to the target side only after the placement or flip is complete (temporal coordination). The correct action for one arm at a given step therefore depends on both the position and the progress of the other~\cite{krebs2024constraints}.

Existing bimanual policies include two common designs~\cite{qicheng2026pabicoop}.
One line of work, combined action policies, uses an action head that combines the left and right arms' actions at the same timestep into one token~\cite{black2024pi0,intelligence2025pi05,reuss2025flower,shukor2025smolvla,yuan2026fastwam,yang2026lila}.
Another line of work, separate action policies, represents each arm individually through dedicated policy branches~\cite{lu2024anybimanual}, twin backbones~\cite{im2025twinvla}, or coordination modules~\cite{lee2025interact,jiang2025dif,motoda2025iace}, which may introduce additional parameters and require more complex network architectures to enable cross-arm communications.
This motivates a better design that enables spatial and temporal interaction between arms.

In this work, we aim to push the frontier of bimanual manipulation by proposing a simple but effective method: \textbf{Duo}, in which one shared action head generates the action chunks~\cite{zhao2023act} of both arms.
Each arm has its own state and action tokens, so self-attention can directly interact across both arms and future timesteps.
As illustrated in \Cref{fig:duovla_comparison}, unlike combined action policies that mix both arms into one token, or separate action designs that generate each arm independently, Duo jointly models spatial and temporal coordination through self-attention over arm-specific action tokens within the shared head during flow-matching generation~\cite{lipman2022flowmatching,liu2023rectifiedflow}.

We integrate Duo into both vision-language-action (VLA) and world-action model (WAM) policies and evaluate it across simulation benchmarks and real-world experiments.
On RoboTwin 2.0~\cite{chen2025robotwin}, Duo improves the overall success of SmolVLA~\cite{shukor2025smolvla}, Lila-WAM~\cite{yang2026lila}, FLOWER~\cite{reuss2025flower}, and $\pi_{0.5}$~\cite{intelligence2025pi05}, covering 0.45B to 3.62B parameters.
Among them, FLOWER-Duo achieves the largest performance gain of 12.55 percentage points, while $\pi_{0.5}$-Duo gains 9.02 points and achieves the highest overall success.
We further evaluate FLOWER-Duo on RoboDojo~\cite{chen2026robodojo} and real-robot tasks.
On the 42-task RoboDojo benchmark, it outperforms StarVLA~\cite{community2026starvla} and Fast-WAM~\cite{yuan2026fastwam}, both with $4-7\times$ more parameters, and achieves the best precision, memory, and randomized-generalization scores among all compared policies.
On a real dual-arm robot, FLOWER-Duo outperforms $\pi_{0.5}$ across three real-world tasks, namely soft-parcel handling, rigid-parcel handling, and box packing.
Meanwhile, we show that FLOWER-Duo scales effectively with increasing training data on the rigid-parcel handling task, outperforming $\pi_{0.5}$ as data grows from 2 to 32 hours.

To provide deeper insight into the design choices of Duo, we examine how the two arms interact during generation and how action dimensions are organized into tokens.
Regarding inter-arm coordination, compared with combined action and separate action baselines, Duo achieves higher success with the same or fewer parameters, whereas masking cross-arm attention at test time substantially lowers success, highlighting the importance of cross-arm attention.
Regarding action organization and granularity, although finer tokenization brings marginal gains, Duo captures nearly all of the improvement at $6.12\times$ lower compute, offering a favorable balance between performance and efficiency.





Overall, our contributions are as follows:
\begin{itemize}[leftmargin=1.5em, itemsep=2pt, topsep=2pt]
    \item We introduce \textbf{Duo}, a simple design that supports spatial and temporal coordination via cross-arm self-attention in one shared action head.
    \item Duo shows consistent gains across VLA and WAM policies on RoboTwin 2.0, improves FLOWER on RoboDojo, and demonstrates strong real-world performance and data scaling ability.
    \item Through systematic analysis, we show that cross-arm attention during generation is essential for coordination, and establish physical arms as compact, natural interaction units that balance accuracy and compute.
\end{itemize}


\input{figure/1_duovla}

%% file: figure/0_comparison.tex
\begin{figure*}[t]
    \centering
    \includegraphics[width=\textwidth]{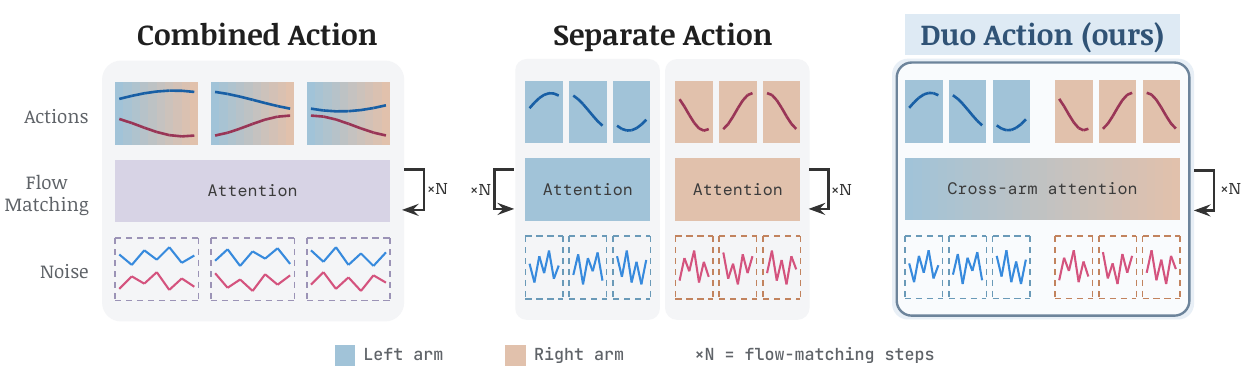}
    \caption{Comparison of bimanual action generation designs.
    Combined action policies combine both arms' actions in each token and use a shared head.
    Separate action policies use arm-specific tokens and independent heads.
    Duo retains arm-specific tokens while jointly generating both arms' actions through cross-arm self-attention in one shared head.
    }
    \label{fig:duovla_comparison}
\end{figure*}


%% file: figure/1_duovla.tex
\begin{figure*}[t]
    \centering

\includegraphics[width=\textwidth]{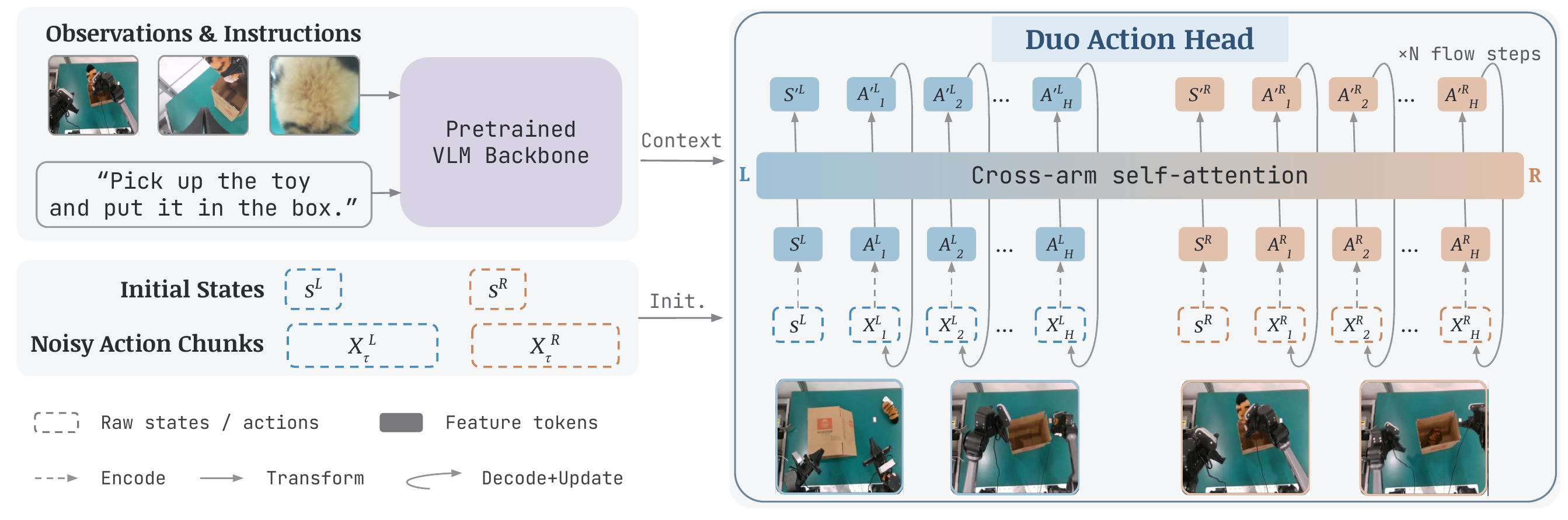}
    \caption{Overview of Duo.
A pretrained VLM backbone encodes the camera images and the language instruction into context for the Duo action head.
The left arm's initial state $s^L$ and noisy actions $X_1^L,\ldots,X_H^L$ are encoded into a state token $S^L$ and action tokens $A_1^L,\ldots,A_H^L$ ({\color{duoblue}blue}), and likewise for the right arm ({\color{duoorange}orange}).
The action head processes the tokens of both arms as one sequence, so cross-arm self-attention lets each token attend to both arms' states and actions.
The output action tokens $A'_h$ are decoded into flows that update the noisy actions, and this is repeated for $N$ flow steps to generate both arms' action chunks jointly.}
    \label{fig:duovla_overview}
\end{figure*}

%% file: sec/2_related_work.tex
\section{Related Work}
\label{sec:related_work}

\subsection{Robotic Policies}
\label{sec:related_vla}

Vision-language-action (VLA) models~\cite{ma2026vlasurvey} and world-action models (WAMs)~\cite{lu2026wamsurvey} represent two prominent directions for transferring pretrained knowledge into robot manipulation policies.
VLAs build on the visual and language understanding of pretrained vision-language models (VLMs) for robot control. Among these, flow-based policies couple these models to action heads that transform noise into continuous action sequences conditioned on visual observations and language instructions~\cite{black2024pi0,reuss2025flower,shukor2025smolvla}.
WAMs further combine understanding with prediction by coupling action generation with the prediction of future observations~\cite{li2026lightwam,wang2026openwam} or latent states~\cite{Dawid_2024, yang2026lila}.
Duo introduces arm-wise action tokens and cross-arm self-attention within the action heads of both VLA and WAM policies.

\subsection{Bimanual Coordination}
\label{sec:related_factorization}

Bimanual policies organize action prediction through joint models~\cite{liu2024rdt,grotz2024peract2}, separate arm branches~\cite{im2025twinvla}, or role-specific decoders~\cite{qicheng2026pabicoop}. Additionally, data-level methods such as CTR reduce spurious cross-arm coupling via temporal recombination~\cite{hu2026ctr}.
Within arm-wise architectures, decomposition enables reusing unimanual skills~\cite{lu2024anybimanual} or dynamically assigning primary and auxiliary roles~\cite{qicheng2026pabicoop}.
To exchange information across arms, prior methods connect arm-specific VLM branches~\cite{im2025twinvla}, exchange conditioning state features~\cite{jiang2025dif}, or introduce decoder synchronization blocks~\cite{lee2025interact} and intermediate coordination encoders~\cite{motoda2025iace}.
In contrast, Duo enables cross-arm interaction directly between action tokens via self-attention in one shared head, without separate branches or auxiliary coordination modules.

\subsection{Action Representations}
\label{sec:related_tokenization}
Action representations determine which action dimensions and timesteps are encoded together~\cite{zhong2025actiontokenization}.
RT-2 and OpenVLA tokenize individual action dimensions~\cite{brohan2023rt2,kim2024openvla}, and FAST compresses information across dimensions and time into discrete token sequences~\cite{pertsch2025fast}.
This grouping choice also arises in continuous policies: $\pi_0$ and SmolVLA embed the full action vector at each timestep into a single token~\cite{black2024pi0,shukor2025smolvla}.
Coordi-$\pi$ separates arm--hand actions into interleaved tokens and adds hand-identity embeddings and factorized positional encoding~\cite{meng2026coordipi}.
With Duo, we study the difference between physical-arm and finer token granularities, examining their effects on coordination, policy performance, and inference computation.

%% file: sec/3_method.tex
\section{Method}
\label{sec:method}

\subsection{Overview}
As shown in \Cref{fig:duovla_overview}, our method builds on flow-based bimanual policies that pair a pretrained backbone with an action head.
Given camera images and a language instruction, the VLM backbone produces context $c$.
Conditioned on $c$ and the initial states $s^L, s^R$ of the left and right arms, the action head predicts a bimanual action chunk $X=[X^L\;X^R]\in\mathbb{R}^{H\times 2d}$ of $H$ future actions, where $X^L, X^R\in\mathbb{R}^{H\times d}$ are the chunks of the two arms.

Duo changes only how this action head represents the two arms, by giving each arm its own tokens, which interact through self-attention within the shared head.

\subsection{Arm-wise Action Tokens}
\label{sec:duovla_tokenization}


Duo encodes each arm into its own tokens.
For one arm, let $s$ be its initial state and $X_h$ the $h$-th step of its noisy action chunk at flow time $\tau$ (\Cref{sec:duovla_flow}). Duo forms the following projection:
\begin{equation}
    S=E_s(s)+e,
    \qquad A_h=E_x(X_h)+e,
    \quad h=1,\ldots,H,
    \label{eq:duovla_arm_tokens}
\end{equation}
where $E_s$ and $E_x$ are the arm's state and action encoders and $e$ is a learned embedding identifying the arm.
Each arm thus contributes one state token and $H$ action tokens (we write $S^L, A_h^L$ and $S^R, A_h^R$ when the arm must be specified)
The tokens of both arms form one sequence of $2(H{+}1)$ tokens in which $A_h^L$ and $A_h^R$ share the temporal position index $h$.

\subsection{Joint Action Generation}
\label{sec:duovla_flow}

Let $A'_h$ denote the head's output for action token $A_h$ of either arm.
Decoding all $A'_h$ yields the predicted flow $v_\theta(X_\tau,\tau;c,s^L,s^R)\in\mathbb{R}^{H\times 2d}$, in which each arm's flow depends on both arms' noisy actions through cross-arm self-attention.
We train $v_\theta$ with rectified flow~\cite{liu2023rectifiedflow,lipman2022flowmatching}.
Given a demonstrated chunk $X$ and noise $\epsilon\sim\mathcal{N}(0,I)$, we form $X_\tau=(1-\tau)\epsilon+\tau X$ with a flow time $\tau\in[0,1]$ shared by both arms, and minimize
\begin{equation}
    \mathcal{L}_{\mathrm{FM}}
    =
    \mathbb{E}_{X,\epsilon,\tau}
    \Big[
        \big\|v_\theta(X_\tau,\tau;c,s^L,s^R)-(X-\epsilon)\big\|_F^2
    \Big].
    \label{eq:duovla_training_objective}
\end{equation}

At inference, starting from $X_\tau=\epsilon$ at $\tau=0$, the head integrates $v_\theta$ to $\tau=1$ over $N$ flow steps (\Cref{fig:duovla_overview}).
Each step recomputes cross-arm attention on the current noisy chunks, allowing both arms to iteratively refine their trajectories in response to each other's evolving motion.

%% file: sec/4_evaluation.tex
\section{Experimental Setup}
\label{sec:evaluation}

To evaluate how Duo improves bimanual policies, we first apply Duo to four VLA and WAM policies on RoboTwin 2.0 and compare them with state-of-the-art open-source baselines, with results presented in \Cref{sec:simulation_results}.

We then evaluate FLOWER-Duo on RoboDojo and test whether its coordination carries over to a real robot, as reported in \Cref{sec:simulation_results,sec:real_results}.

To provide deeper insight into the design of Duo, we isolate Duo's key design choices in \Cref{sec:ablations}.

\input{table/robotwin_main}

\subsection{Benchmarks}
\label{sec:benchmarks}

\paragraph{RoboTwin 2.0.}
RoboTwin 2.0~\cite{chen2025robotwin} comprises 50 bimanual tasks on a simulated Aloha-AgileX robot.
Following the official protocol, we train one multi-task policy on 50 clean demonstrations per task (2,500 total).
The \emph{Clean} setting matches the training domain; the \emph{Random} setting introduces tabletop clutter, random backgrounds and lighting, and table-height perturbations to test out-of-domain (OOD) generalization.
We report success rates in both settings and their mean as the overall score, all in percent.



\paragraph{RoboDojo.}
RoboDojo~\cite{chen2026robodojo} comprises 42 simulated manipulation tasks on a simulated dual-arm ARX X5 robot.
We train on the official set of 3,500 demonstrations from 35 tasks and evaluate all 42 tasks with 50 episodes per task and one seed (2,100 episodes), following the official protocol.
We report its official score, overall success rate (SR), and six capability scores: standard and randomized generalization, precision, long-horizon execution, memory, and open-ended tasks.

\paragraph{Real-world experiments.}
We compare FLOWER-Duo against the strongest simulation baseline, $\pi_{0.5}$, on an AgileX Piper X dual-arm robot across three tasks:

\begin{itemize}
    \item \textbf{Soft-parcel handling.}
    The robot transfers diverse soft parcels from random orientations to an output region label-side up, placing upright ones directly and using one arm to flip and hand over the others. This task tests adaptation to deformation during grasping and handover.
    \item \textbf{Rigid-parcel handling.}
    This task follows the same procedure as soft-parcel handling using rigid parcels, emphasizing precise spatial alignment during handover.
    \item \textbf{Box packing.}
    The robot uses both arms to fold a box and place a plush toy inside, requiring long-horizon coordination and cooperative manipulation of a shared object.
\end{itemize}

\subsection{Baselines}
\label{sec:baselines}

We select eight representative open-source baselines spanning different bimanual designs, and policy families.

By bimanual design, they comprise combined action policies, namely StarVLA~\cite{community2026starvla}, SmolVLA~\cite{shukor2025smolvla}, $\pi_{0.5}$~\cite{intelligence2025pi05}, and FLOWER~\cite{reuss2025flower}; WAMs, namely Fast-WAM~\cite{yuan2026fastwam}, and Lila-WAM~\cite{yang2026lila}; and separate action policies, including TwinVLA~\cite{im2025twinvla} and InterACT~\cite{lee2025interact}.

We adapt Duo to most of the combined action policies. Due to the limited compute resources, we use StarVLA and Fast-WAM as reference baselines.
TwinVLA and InterACT serve as baselines for separate action policies.

By family, they comprise five VLAs, including SmolVLA, FLOWER, $\pi_{0.5}$, StarVLA and TwinVLA; two WAMs, including Lila-WAM, and Fast-WAM; and one  visuomotor bimanual policy InterACT~\cite{lee2025interact}.


%% file: table/robotwin_main.tex
\providecommand{\gain}[1]{{\scriptsize\color{green!45!black}\,+#1}}
\providecommand{\loss}[1]{{\scriptsize\color{red!70!black}\,$-$#1}}

\begin{table}[t]
  \centering
  \caption{Results on RoboTwin 2.0, including Clean, Random, and mean as overall success rate (\%).}
  \label{tab:robotwin_main}
  \definecolor{twinPurple}{HTML}{9886BC}
  \definecolor{twinBlue}{HTML}{65A7D2}
  \definecolor{twinOrange}{HTML}{DFA17B}
  \small
  \setlength{\tabcolsep}{12pt}
  \renewcommand{\arraystretch}{1.15}
  \begin{tabular}{lcccc}
    \toprule
    Policy & Params (B) & Overall & Clean & Random \\
    \midrule
    \rowcolor{black!8}
    \multicolumn{5}{c}{\textit{Separate action policies}} \\
    InterACT & \cellcolor{twinPurple!55}\textbf{0.10} & \cellcolor{twinPurple!9}27.52 & \cellcolor{twinBlue!9}53.34 & \cellcolor{twinOrange!9}1.70 \\
    TwinVLA  & \cellcolor{twinPurple!9}1.30 & \cellcolor{twinPurple!9}25.45 & \cellcolor{twinBlue!9}42.00 & \cellcolor{twinOrange!9}8.90 \\
    \midrule
    \rowcolor{black!8}
    \multicolumn{5}{c}{\textit{Combined action policies \& +Duo}} \\
    SmolVLA & \cellcolor{twinPurple!28}0.45 & \cellcolor{twinPurple!9}22.31 & \cellcolor{twinBlue!9}41.18 & \cellcolor{twinOrange!9}3.44 \\
    \rowcolor{black!5} \quad + Duo &  & \cellcolor{twinPurple!9}24.91\gain{2.60} & \cellcolor{twinBlue!9}46.08\gain{4.90} & \cellcolor{twinOrange!9}3.74\gain{0.30} \\
    \midrule
    Lila-WAM & \cellcolor{twinPurple!9}0.50 & \cellcolor{twinPurple!9}32.91 & \cellcolor{twinBlue!9}58.92 & \cellcolor{twinOrange!9}6.90 \\
    \rowcolor{black!5} \quad + Duo &  & \cellcolor{twinPurple!9}34.01\gain{1.10} & \cellcolor{twinBlue!9}63.12\gain{4.20} & \cellcolor{twinOrange!9}4.90\loss{2.00} \\
    \midrule
    FLOWER & \cellcolor{twinPurple!9}0.89 & \cellcolor{twinPurple!9}50.15 & \cellcolor{twinBlue!9}68.68 & \cellcolor{twinOrange!9}31.62 \\
    \rowcolor{black!5} \quad + Duo &  & \cellcolor{twinPurple!28}62.70\gain{12.55} & \cellcolor{twinBlue!28}78.74\gain{10.06} & \cellcolor{twinOrange!28}46.66\gain{15.04} \\
    \midrule
    $\pi_{0.5}$ & \cellcolor{twinPurple!9}3.62 & \cellcolor{twinPurple!9}58.35 & \cellcolor{twinBlue!9}70.70 & \cellcolor{twinOrange!9}46.00 \\
    \rowcolor{black!5} \quad + Duo &  & \cellcolor{twinPurple!55}\textbf{67.37}\gain{9.02} & \cellcolor{twinBlue!55}\textbf{80.36}\gain{9.66} & \cellcolor{twinOrange!55}\textbf{54.38}\gain{8.38} \\
    \midrule
    StarVLA  & \cellcolor{twinPurple!9}4.50 & \cellcolor{twinPurple!9}24.84 & \cellcolor{twinBlue!9}46.52 & \cellcolor{twinOrange!9}3.16 \\
    Fast-WAM & \cellcolor{twinPurple!9}6.00 & \cellcolor{twinPurple!9}39.85 & \cellcolor{twinBlue!9}77.80 & \cellcolor{twinOrange!9}1.90 \\
    \bottomrule
  \end{tabular}
\end{table}

%% file: sec/5_results.tex
\ifdefined\DuoPreprintMode\else\input{figure/3_group_delta}\fi
\input{table/robodojo_main}

\section{Results}
\label{sec:results}

\subsection{Simulation Benchmarks}
\label{sec:simulation_results}

\paragraph{On RoboTwin2.0, Duo improves policies across architectures and scales.}
As shown in \Cref{tab:robotwin_main}, Duo improves overall success across VLA and WAM policies.
Among them, $\pi_{0.5}$-Duo achieves the highest overall success at 67.37\% and leads in both Clean and Random settings.
Meanwhile, FLOWER-Duo achieves the largest gain of 12.55 points, reaching 62.70\% and surpassing the original $\pi_{0.5}$ with about a quarter of its parameters.

Notably, while all four policies improve in Clean scenes, Random-setting gains concentrate on FLOWER and $\pi_{0.5}$, which gain 15.04 points and 8.38 points, respectively.
Because Duo targets bimanual action modeling, it cannot compensate when a base policy suffers near-total failure under OOD, as seen in SmolVLA (3.44\%) and Lila-WAM (6.90\%).
Once the base policy retains viable OOD performance, Duo's improvements carry over strongly.

We use FLOWER-Duo, which achieves the largest overall gain, to further examine how Duo benefits different forms of arm coordination.

\ifdefined\DuoPreprintMode\input{figure/3_group_delta}\par\fi \paragraph{On RoboTwin2.0, FLOWER-Duo gains most when both arms move together.}
Specifically, we divide tasks into three groups according to the arm motions in the demonstrations: one arm, arms together, and arms in turn.

As illustrated in \Cref{fig:robotwin_group_delta}, FLOWER-Duo achieves the largest gains on arms together tasks, which are 14.3 and 16.5 points in the Clean and Random settings, respectively. The reason possibly is that attention across arms and future timesteps may help align the two arms' predicted motions in space and time during concurrent manipulation. 

It also outperforms $\pi_{0.5}$ by 13.7 points under randomization on arms in turn tasks, while remaining competitive on one arm tasks. In arms in turn tasks such as ``Handover Block'' and ``Handover Mic'', attention across arms and time may help a later action account for the other arm's preceding motion, thus facilitating smoother handovers and tighter inter-arm alignment.


\paragraph{On RoboDojo, FLOWER-Duo leads in precision.}
The results in \Cref{tab:robodojo_main} show that RoboDojo remains challenging, with none of the evaluated policies exceeding 7\% overall success.
The 0.89B FLOWER-Duo ranks second in both score and success rate, behind only $\pi_{0.5}$ and ahead of the larger StarVLA and Fast-WAM policies.
With the same backbone, Duo also improves FLOWER's precision success rate, increasing from 8.83\% to 12.88\%.
Duo allows each arm's predicted actions to condition on the other arm's proposed motions across future timesteps.
This interaction may help align the arms' positions and action timing, reducing coordination errors that can disrupt execution.

Beyond precision, Duo also improves FLOWER's performance across multiple capability dimensions.
These gains suggest that coordination is a shared execution requirement across tasks with different capability demands. Improving how the arms act together can help the policy complete tasks whose challenges also involve scene variation, extended action sequences, and memory.

\input{figure/4_real_robot}

\subsection{Real-world Experiments}
\label{sec:real_results}

We compare FLOWER-Duo with $\pi_{0.5}$ as our baseline, on three real-world tasks that require temporal alignment, spatial coordination and complex long-horizon execution.

\input{table/real_robot}

\paragraph{FLOWER-Duo outperforms $\pi_{0.5}$ across all three real-world experiments.}
As reported in \Cref{tab:real_robot}, FLOWER-Duo achieves 90.0\%, 83.8\%, and 40.0\% success, respectively, surpassing $\pi_{0.5}$ in all tasks.
It also achieves higher success on out-of-domain parcels, with OOD gains of 20.0 and 5.0 percentage points on soft and rigid-parcel handling, respectively.
Its advantage extends from sequential handovers to simultaneous manipulation, although box packing, which is a longer and more complex task, remains challenging for both policies.

To illustrate the execution differences between the two policies, we provide qualitative examples of soft-parcel handling and box packing in \Cref{fig:real_robot_rollouts}. 
In the soft-parcel example, $\pi_{0.5}$ fails at the flipping stage, whereas FLOWER-Duo successfully flips the parcel and places it label-side up.
In the box-packing example, $\pi_{0.5}$ overturns the box during stabilization, whereas FLOWER-Duo keeps it stable and places the toy inside successfully.

\input{figure/6_real_robot_scaling}

\paragraph{FLOWER-Duo completes successful trials faster.}
Across all three real-robot tasks, FLOWER-Duo completes successful trials faster than $\pi_{0.5}$ on average.
One possible explanation is that cross-arm self attention allows each arm's predicted trajectory to account for the other arm's future motion.
During handling, this may help the receiving arm prepare for the approaching parcel; during box packing, it may help both grippers coordinate contact and motion.
Such coordination could reduce waiting and corrective movements.

\paragraph{FLOWER-Duo scales with training data.}

Having established FLOWER-Duo's advantages under the full training budget, we next investigate how its success rate and completion time scale with increasing training data.


As shown in~\Cref{fig:real_robot_scaling}, while $\pi_{0.5}$ holds an early edge under very small budgets, FLOWER-Duo scales more strongly as data grows, matching $\pi_{0.5}$ at 8 hours and surpassing it at both 16 and 32 hours without saturating.
This demonstrates that by explicitly modeling inter-arm dependencies through arm-wise tokens and cross-arm attention, FLOWER-Duo learns coordination more efficiently, outscaling the much larger $\pi_{0.5}$ as training data increase.

Beyond success rate, broader demonstration coverage may also improve execution efficiency, bringing completion times closer to human expert demonstrations\footnote{The mean human demonstration time is 18.3\,s, computed from 20 demonstrations randomly sampled from the real-robot training dataset.} as hesitations and corrective motions decrease.
This echoes Hugging Face's finding that fine-tuning $\pi_{0.5}$ on curated, consistent demonstrations yields faster and smoother task execution in cloth folding~\cite{kooijmans2026unfolding}.

%% file: figure/3_group_delta.tex
\ifdefined\DuoPreprintMode
\begin{wrapfigure}{r}{0.58\textwidth}
\centering
\vspace{-6pt}
\includegraphics[width=\linewidth]{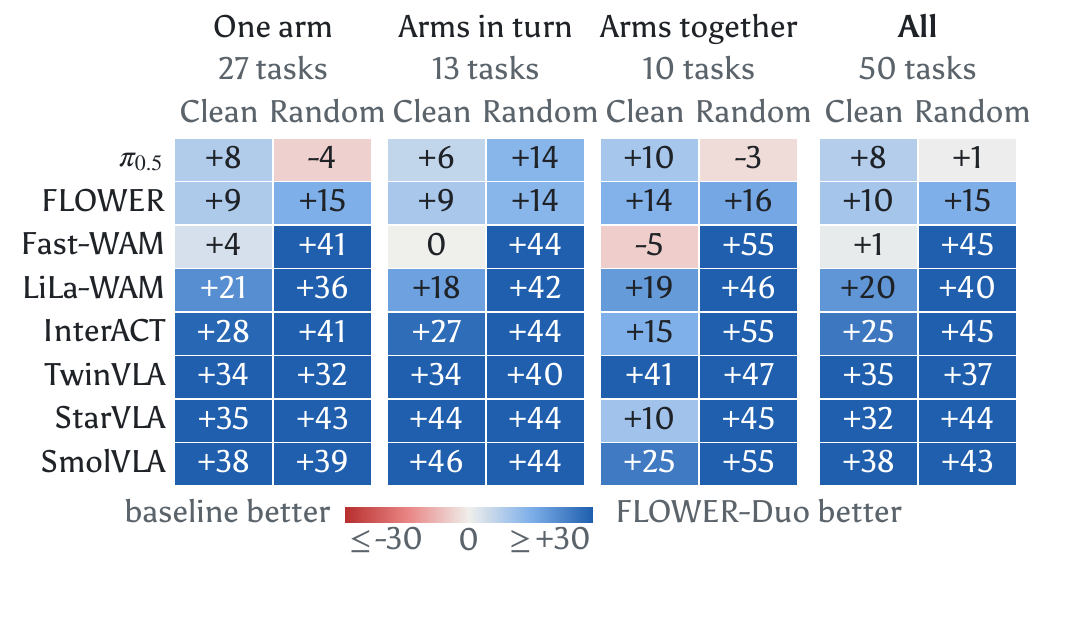}
\caption{FLOWER-Duo success-rate gains (percentage points) on RoboTwin 2.0.}
\label{fig:robotwin_group_delta}
\vspace{-8pt}
\end{wrapfigure}
\else
\begin{figure}[!t]
\centering
\includegraphics[width=\columnwidth]{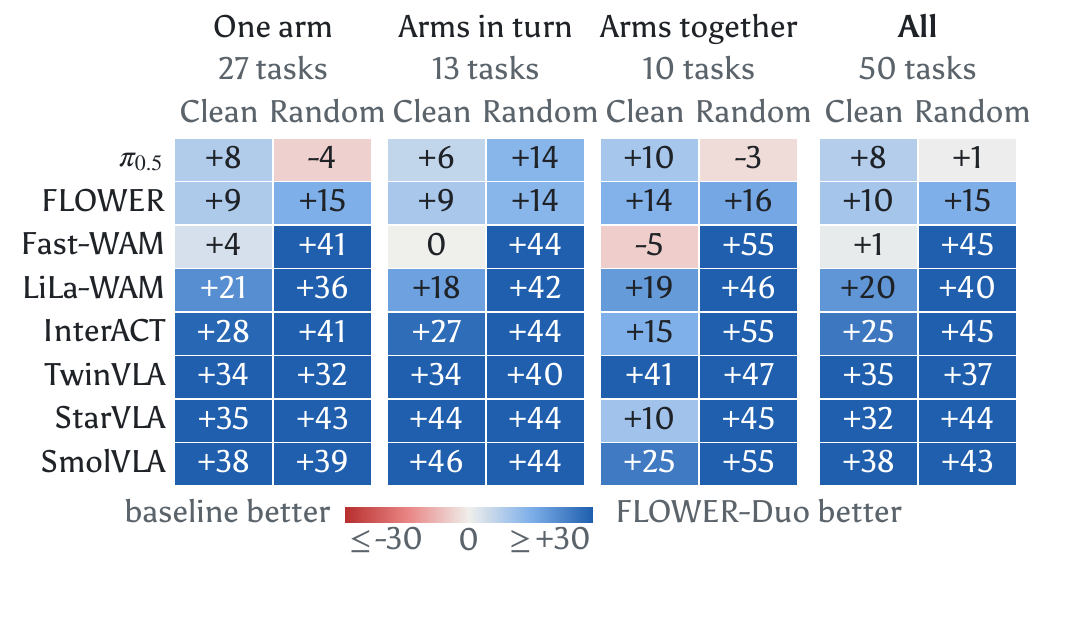}
\vspace{-8pt}
\caption{FLOWER-Duo success-rate gains (percentage points) on RoboTwin 2.0.}
\label{fig:robotwin_group_delta}
\end{figure}
\fi

%% file: table/robodojo_main.tex
\begin{table*}[t]
  \caption{Results on RoboDojo, including score, success rate (\%,SR), and six capability metrics. }
  \label{tab:robodojo_main}
  \centering
  \definecolor{dojoPurple}{HTML}{9886BC}
  \definecolor{dojoOrange}{HTML}{DFA17B}
  \definecolor{dojoBlue}{HTML}{65A7D2}
  \footnotesize
  \setlength{\tabcolsep}{3.2pt}
  \renewcommand{\arraystretch}{1.14}
  \begin{tabular*}{\textwidth}{@{\extracolsep{\fill}}l*{9}{c}}
    \toprule
    Method & Params (B) & Score & SR (\%) & Gen-Std & Gen-Rand & Precision & Long & Memory & Open \\
    \midrule
    \rowcolor{black!8}
    \multicolumn{10}{c}{\textit{Combined action policies}} \\
    Fast-WAM                   & \cellcolor{dojoPurple!9}6.00 & \cellcolor{dojoPurple!9}3.48 & \cellcolor{dojoPurple!9}2.03 & \cellcolor{dojoBlue!9}2.00 & \cellcolor{dojoBlue!9}0.00 & \cellcolor{dojoOrange!9}0.00 & \cellcolor{dojoOrange!9}5.17 & \cellcolor{dojoOrange!9}3.44 & \cellcolor{dojoOrange!9}0.42 \\
    StarVLA                    & \cellcolor{dojoPurple!9}4.50 & \cellcolor{dojoPurple!9}6.40 & \cellcolor{dojoPurple!9}3.24 & \cellcolor{dojoBlue!9}5.00 & \cellcolor{dojoBlue!9}0.00 & \cellcolor{dojoOrange!9}4.33 & \cellcolor{dojoOrange!9}6.50 & \cellcolor{dojoOrange!9}2.44 & \cellcolor{dojoOrange!9}0.58 \\
    $\pi_{0.5}$                & \cellcolor{dojoPurple!9}3.62 & \cellcolor{dojoPurple!55}\textbf{11.41} & \cellcolor{dojoPurple!55}\textbf{6.91} & \cellcolor{dojoBlue!55}\textbf{15.00} & \cellcolor{dojoBlue!28}1.00 & \cellcolor{dojoOrange!9}5.50 & \cellcolor{dojoOrange!55}\textbf{14.67} & \cellcolor{dojoOrange!28}4.56 & \cellcolor{dojoOrange!55}\textbf{1.67} \\
    FLOWER & \cellcolor{dojoPurple!9}0.89 & \cellcolor{dojoPurple!9}4.89 & \cellcolor{dojoPurple!9}1.63 & \cellcolor{dojoBlue!9}3.98 & \cellcolor{dojoBlue!9}0.73 & \cellcolor{dojoOrange!28}8.83 & \cellcolor{dojoOrange!9}8.75 & \cellcolor{dojoOrange!9}3.00 & \cellcolor{dojoOrange!28}1.50 \\
    SmolVLA                   & \cellcolor{dojoPurple!28}0.45 & \cellcolor{dojoPurple!9}0.38 & \cellcolor{dojoPurple!9}0.03 & \cellcolor{dojoBlue!9}0.33 & \cellcolor{dojoBlue!9}0.00 & \cellcolor{dojoOrange!9}0.00 & \cellcolor{dojoOrange!9}0.00 & \cellcolor{dojoOrange!9}0.00 & \cellcolor{dojoOrange!9}0.00 \\
    \midrule[0.4pt]
    \rowcolor{black!8}
    \multicolumn{10}{c}{\textit{Separate action policies}} \\
    TwinVLA        & \cellcolor{dojoPurple!9}1.30 & \cellcolor{dojoPurple!9}0.70 & \cellcolor{dojoPurple!9}0.27 & \cellcolor{dojoBlue!9}1.33 & \cellcolor{dojoBlue!9}0.00 & \cellcolor{dojoOrange!9}0.00 & \cellcolor{dojoOrange!9}0.00 & \cellcolor{dojoOrange!9}0.67 & \cellcolor{dojoOrange!9}0.00 \\
    InterACT & \cellcolor{dojoPurple!55}\textbf{0.10} & \cellcolor{dojoPurple!9}0.61 & \cellcolor{dojoPurple!9}0.18 & \cellcolor{dojoBlue!9}0.75 & \cellcolor{dojoBlue!9}0.00 & \cellcolor{dojoOrange!9}0.75 & \cellcolor{dojoOrange!9}1.31 & \cellcolor{dojoOrange!9}0.72 & \cellcolor{dojoOrange!9}0.00 \\
    \midrule
    \textbf{FLOWER-Duo}            & \cellcolor{dojoPurple!9}0.89 & \cellcolor{dojoPurple!28}7.07 & \cellcolor{dojoPurple!28}3.75 & \cellcolor{dojoBlue!28}11.43 & \cellcolor{dojoBlue!55}\textbf{2.27} & \cellcolor{dojoOrange!55}\textbf{12.88} & \cellcolor{dojoOrange!28}9.84 & \cellcolor{dojoOrange!55}\textbf{5.52} & \cellcolor{dojoOrange!9}0.25 \\
    \bottomrule
  \end{tabular*}
  \par\vspace{2pt}
\end{table*}

%% file: figure/4_real_robot.tex
\begin{figure*}[t]
    \centering
    \includegraphics[width=\textwidth]{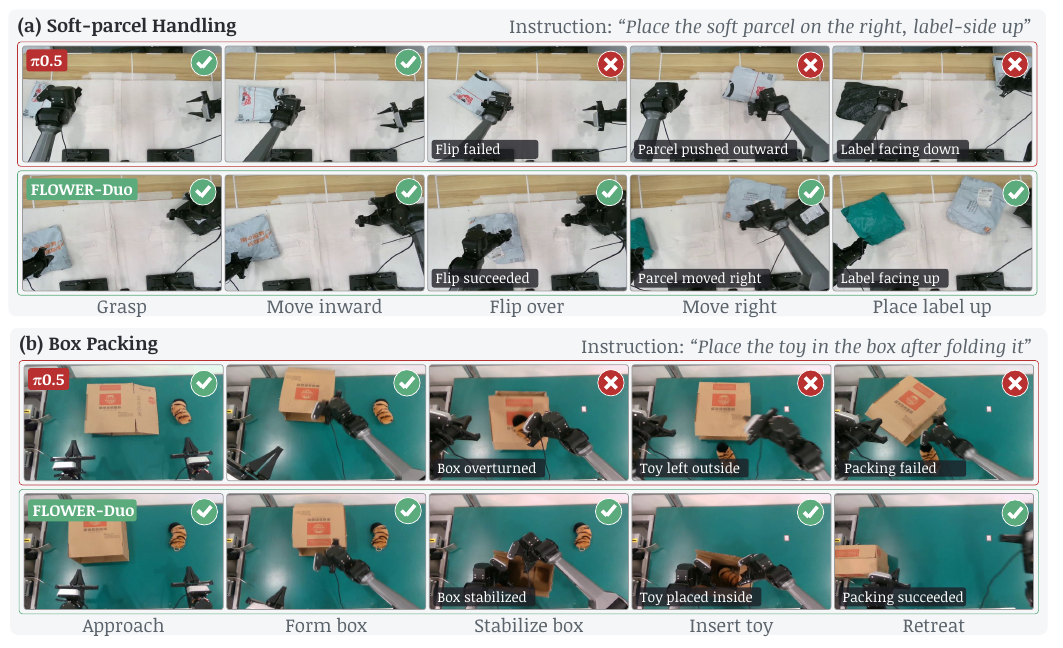}
    \caption{Examples for real-world rollouts of FLOWER-Duo and $\pi_{0.5}$ on soft-parcel handling and box packing.}
    \label{fig:real_robot_rollouts}
\end{figure*}

%% file: table/real_robot.tex
\ifdefined\DuoPreprintMode
\begin{wraptable}{r}{0.55\textwidth}
\vspace{-6pt}
\caption{Real-robot success rates (\%) and mean completion times for successful trials ($T$, seconds).}
\label{tab:real_robot}
\centering
\scriptsize
\setlength{\tabcolsep}{1.8pt}
\renewcommand{\arraystretch}{1.15}
\resizebox{\linewidth}{!}{%
\begin{tabular}{@{}lccccccccccc@{}}
\toprule
& \multicolumn{4}{c}{\shortstack{Soft-parcel\\handling}} & \multicolumn{4}{c}{\shortstack{Rigid-parcel\\handling}} & \multicolumn{2}{c}{\shortstack{Box\\packing}} \\
\cmidrule(lr){2-5} \cmidrule(lr){6-9} \cmidrule(l){10-11}
Model & ID & OOD & All & $T$(s) & ID & OOD & All & $T$(s) & SR & $T$(s) \\
\midrule
$\pi_{0.5}$ & 82.5 & 65.0 & 73.8 & 25.5 & 67.5 & 77.5 & 72.5 & 34.8 & 20.0 & 28.8 \\
\textbf{FLOWER-Duo} & \textbf{95.0} & \textbf{85.0} & \textbf{90.0} & \textbf{22.7} & \textbf{85.0} & \textbf{82.5} & \textbf{83.8} & \textbf{33.1} & \textbf{40.0} & \textbf{27.1} \\
\bottomrule
\end{tabular}%
}
\vspace{-6pt}
\end{wraptable}
\else
\begin{table}[t]
\caption{Real-robot success rates (\%) and mean completion times for successful trials ($T$, seconds).}
\label{tab:real_robot}
\centering
\footnotesize
\setlength{\tabcolsep}{1.15pt}
\renewcommand{\arraystretch}{1.15}
\begin{tabular*}{\columnwidth}{@{\extracolsep{\fill}}lccccccccccc@{}}
\toprule
& \multicolumn{4}{c}{\shortstack{Soft-parcel\\handling}} & \multicolumn{4}{c}{\shortstack{Rigid-parcel\\handling}} & \multicolumn{2}{c}{\shortstack{Box\\packing}} \\
\cmidrule(lr){2-5} \cmidrule(lr){6-9} \cmidrule(l){10-11}
Model & ID & OOD & All & $T$(s) & ID & OOD & All & $T$(s) & SR & $T$(s) \\
\midrule
$\pi_{0.5}$ & 82.5 & 65.0 & 73.8 & 25.5 & 67.5 & 77.5 & 72.5 & 34.8 & 20.0 & 90.0 \\
\textbf{FLOWER-Duo} & \textbf{95.0} & \textbf{85.0} & \textbf{90.0} & \textbf{22.7} & \textbf{85.0} & \textbf{82.5} & \textbf{83.8} & \textbf{33.1} & \textbf{40.0} & \textbf{33.6} \\
\bottomrule
\end{tabular*}
\end{table}
\fi

%% file: figure/6_real_robot_scaling.tex
\ifdefined\DuoPreprintMode
\begin{wrapfigure}{r}{0.5\textwidth}
\centering
\vspace{-6pt}
\includegraphics[width=\linewidth]{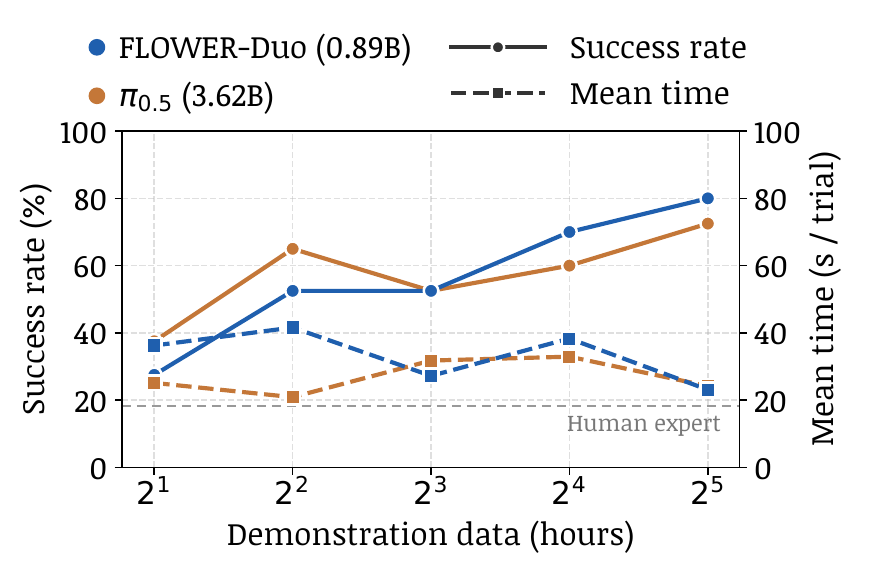}
\caption{Success rate (\%) and mean execution time over successful trials with increasing training data on rigid-parcel handling. The gray dashed line shows the human demonstration mean time.}
\label{fig:real_robot_scaling}
\vspace{-6pt}
\end{wrapfigure}
\else
\begin{figure}[t]
\centering
\includegraphics[width=\columnwidth]{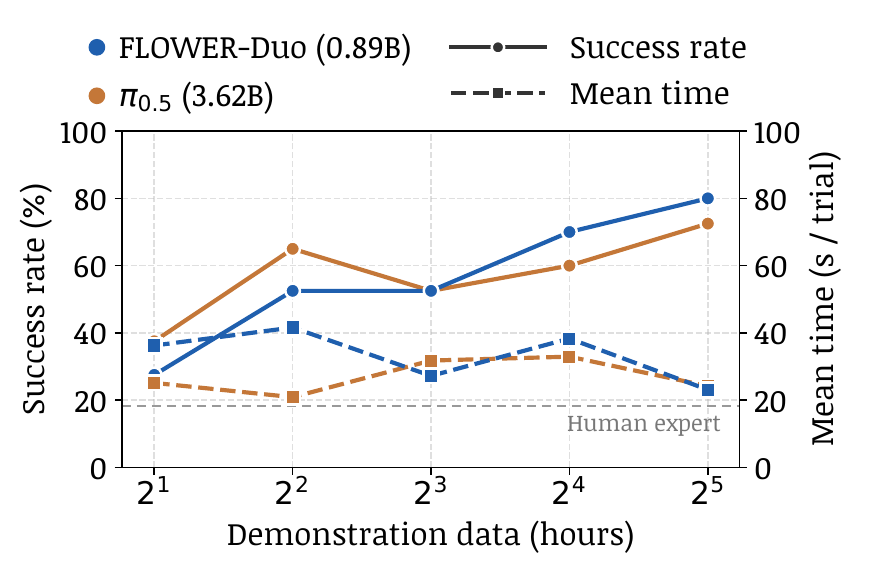}
\caption{Success rate (\%) and mean execution time over successful trials with increasing training data on rigid-parcel handling. The gray dashed line shows the human demonstration mean time.}
\label{fig:real_robot_scaling}
\end{figure}
\fi

%% file: sec/6_ablations.tex
\section{Ablation Studies}
\label{sec:ablations}

We systematically examine how inter-arm coordination, action grouping, and token granularity affect policy performance through the following two questions:
\begin{enumerate}[label=\textbf{Q\arabic*.}, leftmargin=2.3em, labelsep=0.5em, itemsep=2pt, topsep=3pt, parsep=0pt]
    \item \emph{Does cross-arm interaction improve performance beyond separating arm representations?} 
    
    We compare generating the two arms' actions separately versus jointly, and test whether joint generation benefits from cross-arm attention in \Cref{sec:coupling_ablation}.
    \item \emph{Why use one token per physical arm at each timestep?} 
    We explore different action token groupings and token granularity to address this question in  \Cref{sec:action_partition_ablation}.
\end{enumerate}

\subsection{Inter-arm coordination}
\label{sec:coupling_ablation}

\ifdefined\DuoPreprintMode\else\input{table/coupling_ablation}\fi

To examine the role of cross-arm interaction in policies with separate arm representations (Q1), we conduct the comparisons using the same backbone:
\begin{itemize}
    \item \textbf{Combined action.}
    Both arms' actions are embedded into one token per horizon position and generated by one action head, as in most VLA policies~\cite{black2024pi0,intelligence2025pi05,reuss2025flower}.
    \item \textbf{Separate action.}
    Each arm has its own tokens and action head, with shared observation context but no access to the other arm's action tokens.
    \item \textbf{FLOWER-Duo.}
    Each arm has its own tokens, and both arms are generated by one shared action head in which each arm also attends to the other arm's noisy actions.
\end{itemize}

\ifdefined\DuoPreprintMode\input{table/coupling_ablation}\par\fi \paragraph{FLOWER-Duo outperforms separate action with fewer parameters.}
The separate action baseline improves overall success from 50.15\% to 60.30\%, while FLOWER-Duo reaches 62.70\%, outperforming it in both Clean and Random settings with fewer parameters (0.89B versus 1.07B).
Although the two action heads receive the same observations, neither can condition its predictions on the other’s proposed actions.
FLOWER-Duo enables this interaction during denoising, allowing each arm to refine its trajectory in relation to the other's motion.
By coupling these updates, Duo may better align the arms' relative positions and timing while retaining a distinct action sequence for each arm.

\paragraph{Blocking cross-arm attention substantially reduces success.}
Masking cross-arm attention in FLOWER-Duo at test time reduces overall success by 8.10 points to 54.60\%, with drops in both settings, even falling below the separate action baseline.
This difference may arise because FLOWER-Duo learns to use the other arm's proposed motions during training.
Removing that information at inference disrupts the interaction on which the trained policy relies, supporting the role of cross-arm attention in Duo's performance.

\ifdefined\DuoPreprintMode\else\input{figure/5_token_granularity}\fi

\input{figure/7_token_grouping_attention}

\subsection{Token Grouping and Granularity}
\label{sec:action_partition_ablation}

To examine why one token per physical arm is an effective choice (Q2), we vary how action dimensions are grouped into tokens and the granularity at which they interact through attention. We compare five configurations spanning four token counts:

\ifdefined\DuoPreprintMode\input{table/action_partition_ablation}\par\fi \paragraph{Splitting the joint token improves success.}
As shown in \Cref{tab:action_partition_ablation}, all multi-token groupings outperform the joint-token baseline in both Clean and Random settings.

However, how the dimensions are grouped also matters. Among the two-token representations, \emph{Physical arms (Duo)} achieves the highest success in both settings, outperforming \emph{Joint-gripper} by 2.34 percentage points in the Clean setting and 2.78 points in the Random setting.

These results suggest that splitting the action is useful, but preserving arm structure provides a further advantage.
Separate tokens allow actions to attend to relevant context individually; grouping each arm's joints and gripper together may further help the policy learn their coupled motion, and attention between arms supports coordination.

\ifdefined\DuoPreprintMode\input{figure/5_token_granularity}\par\fi \paragraph{Finer tokenization offers limited gains at substantially higher computation.}
We compare action-head GFLOPs per action chunk, normalized to FLOWER-Duo in \Cref{fig:token_granularity}.
Moving from one joint token to Duo captures most of the performance gain.
Further splitting each arm's joints and gripper provides no additional gain despite requiring $1.80\times$ the action-head computation.
Using one token per dimension improves overall success by only 0.78 percentage points over Duo at $6.12\times$ the computation.

This limited return suggests that beyond cross-arm interaction, attention between individual dimensions adds little, while processing the longer token sequence incurs substantial computational and training cost.
Duo therefore provides a favorable balance between performance and computation.

\ifdefined\DuoPreprintMode\else\input{table/action_partition_ablation}\fi

\paragraph{Case study: Attention patterns under different token groupings.}
To better understand the performance differences above, we examine how token grouping organizes attention between action components.
We visualize physical-arm, within-arm, and per-dimension tokenizations on the same roller-grasp observation in \Cref{fig:token_grouping_attention}, using 2, 4, and 14 tokens per timestep, respectively. 

With two arm tokens, both within-arm and cross-arm attention are present.
With four tokens, the joint groups retain strong within-group attention across the horizon, while the gripper groups also attend to one another across arms.
Even after every dimension receives its own token, many of the stronger connections between different dimensions remain within the same physical arm, forming visible arm-aligned blocks along the diagonal of the attention map.

Thus, finer tokenization reveals dependencies that still follow physical-arm boundaries, supporting the physical arm as a meaningful grouping unit.
Duo encodes this structure directly. Each token represents an arm's joints and gripper together, allowing cross-arm attention to exchange information between arm-level actions.

%% file: table/coupling_ablation.tex
\ifdefined\DuoPreprintMode
\begin{wraptable}{r}{0.55\textwidth}
  \vspace{-4pt}
  \caption{Inter-arm coordination ablation on RoboTwin 2.0.}
  \label{tab:coupling_ablation}
  \centering
  \footnotesize
  \setlength{\tabcolsep}{2.5pt}
  \renewcommand{\arraystretch}{1.08}
  \begin{tabular*}{\linewidth}{@{\extracolsep{\fill}}lcccc@{}}
    \toprule
    Action head & Params (B) & Overall & Clean & Random \\
    \midrule
    Combined action & 0.89 & 50.15 & 68.68 & 31.62 \\
    Separate action & 1.07 & 60.30 & 76.56 & 44.04 \\
    \textbf{FLOWER-Duo} & 0.89 & \textbf{62.70} & \textbf{78.74} & \textbf{46.66} \\
    \quad w/o cross-arm attn. & 0.89 & 54.60 & 69.60 & 39.60 \\
    \bottomrule
  \end{tabular*}
  \vspace{-4pt}
\end{wraptable}
\else
\begin{table}[t]
  \caption{Inter-arm coordination ablation on RoboTwin 2.0.}
  \label{tab:coupling_ablation}
  \centering
  \footnotesize
  \setlength{\tabcolsep}{3pt}
  \renewcommand{\arraystretch}{1.08}
  \begin{tabular*}{\columnwidth}{@{\extracolsep{\fill}}lcccc@{}}
    \toprule
    Action head & Params (B) & Overall & Clean & Random \\
    \midrule
    Combined action & 0.89 & 50.15 & 68.68 & 31.62 \\
    Separate action & 1.07 & 60.30 & 76.56 & 44.04 \\
    \textbf{FLOWER-Duo} & 0.89 & \textbf{62.70} & \textbf{78.74} & \textbf{46.66} \\
    \quad w/o cross-arm attn. & 0.89 & 54.60 & 69.60 & 39.60 \\
    \bottomrule
  \end{tabular*}
\end{table}
\fi

%% file: figure/5_token_granularity.tex
\ifdefined\DuoPreprintMode
\begin{wrapfigure}{r}{0.45\textwidth}
    \centering
    \vspace{-6pt}
    \includegraphics[width=\linewidth]{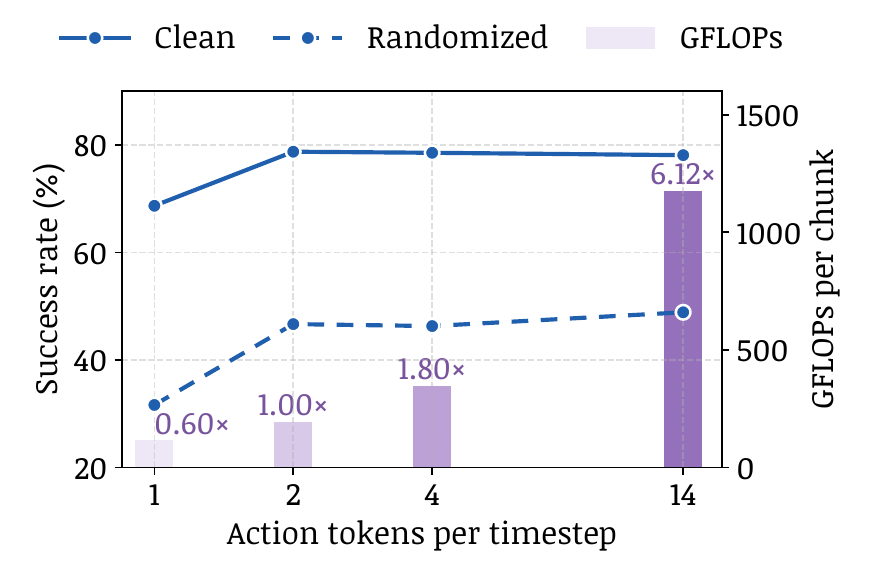}
    \caption{Action-token granularity, success rate(\%), and inference computation on RoboTwin 2.0.}
    \label{fig:token_granularity}
    \vspace{-6pt}
\end{wrapfigure}
\else
\begin{figure}[t]
    \centering
    \includegraphics[width=\columnwidth]{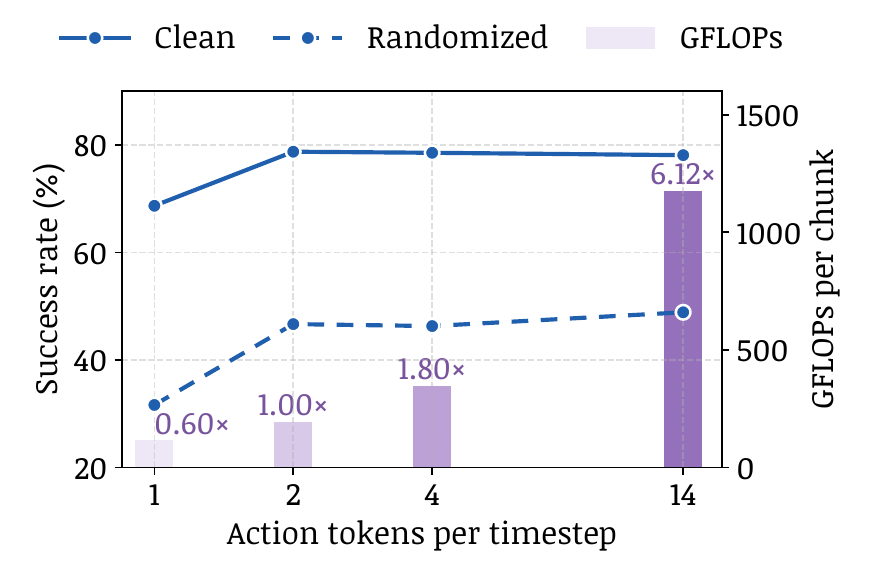}
    \caption{Action-token granularity, success rate(\%), and inference computation on RoboTwin 2.0.}
    \label{fig:token_granularity}
\end{figure}
\fi

%% file: figure/7_token_grouping_attention.tex
\begin{figure*}[t]
    \centering
    \includegraphics[width=\textwidth]{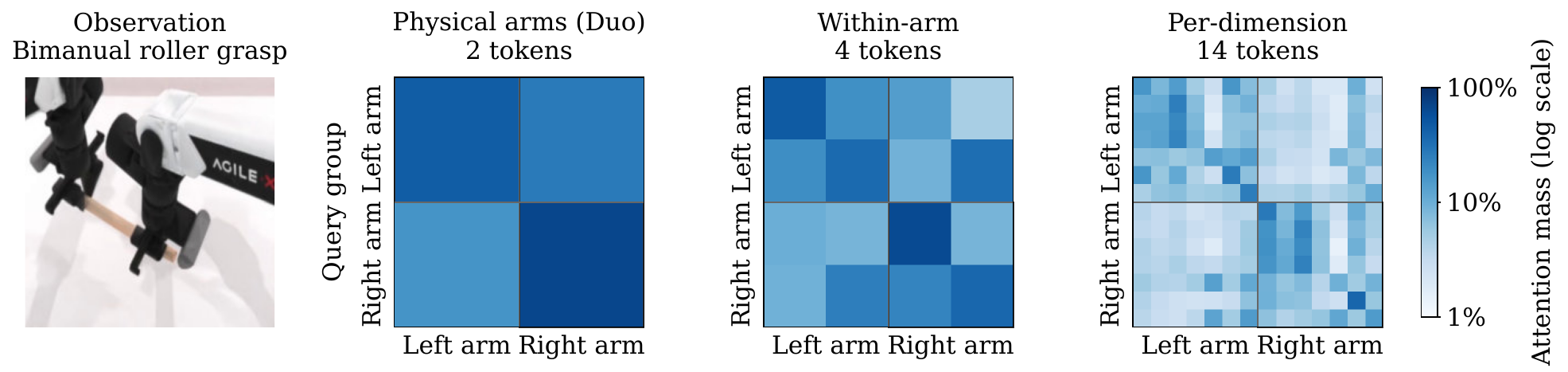}
    \caption{Action attention maps under physical-arm, within-arm, and per-dimension groupings for the same bimanual roller-grasp observation. All maps share a logarithmic color scale.}
    \label{fig:token_grouping_attention}
\end{figure*}

%% file: table/action_partition_ablation.tex

\ifdefined\DuoPreprintMode
\begin{wraptable}{r}{0.50\textwidth}
  \vspace{-6pt}
  \caption{Action grouping and token granularity on RoboTwin 2.0.}
  \label{tab:action_partition_ablation}
  \centering
  \footnotesize
  \setlength{\tabcolsep}{3pt}
  \renewcommand{\arraystretch}{1.08}
  \begin{tabular*}{\linewidth}{@{\extracolsep{\fill}}lcccc@{}}
    \toprule
    Grouping & Dimensions & Overall & Clean & Random \\
    \midrule
    Joint token & 14 & 50.15 & 68.68 & 31.62 \\
    Joint-gripper & 12+2 & 60.14 & 76.40 & 43.88 \\
    \textbf{Physical arms (Duo)} & 7+7 & 62.70 & \textbf{78.74} & 46.66 \\
    Within-arm & 6+6+1+1 & 62.42 & 78.54 & 46.30 \\
    Per-dimension & $14\times1$ & \textbf{63.48} & 78.10 & \textbf{48.86} \\
    \bottomrule
  \end{tabular*}
  \vspace{-6pt}
\end{wraptable}
\else
\begin{table}[t]
  \caption{Action grouping and token granularity on RoboTwin 2.0.}
  \label{tab:action_partition_ablation}
  \centering
  \footnotesize
  \setlength{\tabcolsep}{4pt}
  \renewcommand{\arraystretch}{1.1}
  \begin{tabular*}{\columnwidth}{@{\extracolsep{\fill}}lcccc@{}}
    \toprule
    Grouping & Dimensions & Overall & Clean & Random \\
    \midrule
    Joint token & 14 & 50.15 & 68.68 & 31.62 \\
    Joint-gripper & 12+2 & 60.14 & 76.40 & 43.88 \\
    \textbf{Physical arms (Duo)} & 7+7 & 62.70 & \textbf{78.74} & 46.66 \\
    Within-arm & 6+6+1+1 & 62.42 & 78.54 & 46.30 \\
    Per-dimension & $14\times1$ & \textbf{63.48} & 78.10 & \textbf{48.86} \\
    \bottomrule
  \end{tabular*}
\end{table}
\fi

%% file: sec/7_conclusion.tex
\vspace{10pt}
\section{Conclusion}
\label{sec:conclusion}

We presented Duo, a simple design that represents each arm separately and couples their action generation through self-attention in one shared head.
Duo improves overall success across four VLA and WAM policies on RoboTwin 2.0.
FLOWER-Duo also improves performance on RoboDojo and outperforms $\pi_{0.5}$ on three real-robot tasks, with higher success on unseen parcels and faster completion of successful trials.
Our ablations show that cross-arm interaction and physical-arm grouping both contribute to performance, while finer tokenization offers limited gains at substantially higher computation.
Together, these findings support physical arms as effective interaction units for learning coordinated motions within a shared policy.

%% file: sec/8_future.tex
\vspace{15pt}
\section{Limitations}
\label{sec:future}
Although Duo improves bimanual coordination during action generation, several limitations remain. 

First, due to limited compute resources, we have not yet evaluated Duo on larger-scale policies, which we leave for future work.
Second, beyond action modeling, tighter integration between visual and action tokens will be essential for stronger coordination in bimanual policies.
Third, separating each timestep into arm-wise tokens doubles the action-token sequence length, increasing attention computation at inference compared with joint-token baselines.

%% file: sec/X_suppl.tex
\clearpage
\maketitlesupplementary
\appendix


\section{Task Types on RoboTwin 2.0}
\label{sec:task_types}

We group the 50 RoboTwin 2.0 tasks by how the arms move in their 50 clean demonstrations.

\begin{itemize}
    \item \textbf{One arm} (27 tasks): Adjust Bottle, Beat Block Hammer, Click Alarmclock, Click Bell, Dump Bin Bigbin, Move Can Pot, Move Pillbottle Pad, Move Playingcard Away, Move Stapler Pad, Open Laptop, Open Microwave, Place A2B Left, Place A2B Right, Place Container Plate, Place Empty Cup, Place Fan, Place Mouse Pad, Place Object Scale, Place Object Stand, Place Phone Stand, Place Shoe, Press Stapler, Rotate QRcode, Shake Bottle, Shake Bottle Horizontally, Stamp Seal, Turn Switch.
    \item \textbf{Arms in turn} (13 tasks): Blocks Ranking RGB, Blocks Ranking Size, Handover Block, Handover Mic, Hanging Mug, Place Can Basket, Place Object Basket, Put Bottles Dustbin, Put Object Cabinet, Stack Blocks Three, Stack Blocks Two, Stack Bowls Three, Stack Bowls Two.
    \item \textbf{Arms together} (10 tasks): Grab Roller, Lift Pot, Pick Diverse Bottles, Pick Dual Bottles, Place Bread Basket, Place Bread Skillet, Place Burger Fries, Place Cans Plasticbox, Place Dual Shoes, Scan Object.
\end{itemize}

\section{Training Details}
\label{sec:implementation}
\input{table/app_robotwin_hyp}

\subsection{RoboTwin 2.0}

We train on the 2,500 clean demonstrations under the model-specific schedules summarized in \Cref{tab:robotwin_training}.
The training step budget for each model trained in our experiments is configured to ensure convergence, and we evaluate the final checkpoint.

We fine-tune the released SmolVLA base policy using the LeRobot trainer. 
For FLOWER, we initialize the backbone with Florence-2-large weights and the action head with weights from the released FLOWER checkpoint, then fine-tune the policy.
For Lila-WAM, we keep the pretrained DINOv3-L vision encoder frozen, and train the action model from random initialization.
We reproduce InterACT with an ACT-based policy, initializing the visual backbone from pretrained ResNet-18 weights and training the arm-specific action decoders from random initialization.
Each Duo variant is trained using the same configuration as its corresponding base policy, differing only in the use of the Duo action head.

For externally reported baselines, the $\pi_{0.5}$, StarVLA, and Fast-WAM scores are taken from the \href{https://robotwin-platform.github.io/leaderboard}{official RoboTwin 2.0 leaderboard}.
For TwinVLA, we report the aggregate scores from Table~10 of the original paper (42.0\% on Clean/Easy and 8.9\% on Randomized/Hard), which come from a separate policy fine-tuned for each task.

For ablations, we train the ablation variants using the same configuration as FLOWER-Duo on RoboTwin 2.0.

\subsection{RoboDojo}

On RoboDojo, we train on 3,500 official demonstrations recorded at 25\,Hz, using the schedules in \Cref{tab:robodojo_training}.

As on RoboTwin 2.0, we use the released SmolVLA and FLOWER weights and our ACT-based implementation of InterACT.
We freeze the SmolVLA vision-language backbone and fine-tune the action expert.
We train FLOWER and FLOWER-Duo with the same schedule and evaluate the final checkpoint for both. For InterACT, we likewise evaluate its final checkpoint.
For TwinVLA, we initialize from the released pretrained SingleVLA checkpoint following the official training procedure and train a single multi-task bimanual policy across all RoboDojo tasks.

The $\pi_{0.5}$, StarVLA, and Fast-WAM scores come from the \href{https://robodojo-benchmark.com/LeaderBoard}{official RoboDojo leaderboard}.

\subsection{Real-world Experiments}

For three real-world tasks (soft-parcel handling, rigid-parcel handling, and box packing), we train on 9.19 hours, 10.02 hours, and 4.71 hours of teleoperated demonstrations at 30\,Hz.

\Cref{tab:real_robot_training} summarizes the training configurations of the models evaluated on the real-world experiments.
Given the limited amount of demonstration data, we use training loss alone to determine when to stop. On each real-robot task, we train each model until its training loss shows no improvement for 50 consecutive optimizer updates.

\section{Additional Qualitative Real-world Example}

\label{sec:real_robot_details}

\paragraph{Rigid-parcel handling.}
As shown in \Cref{fig:app_real_robot}, both policies initially turn the parcel to expose its label.
During the subsequent transfer, however, the parcel flips back under $\pi_{0.5}$, resulting in a label-down placement.
FLOWER-Duo preserves the label-up orientation while moving the parcel to the target region.



\input{table/app_robodojo_hyp}
\input{table/app_real_hyp}
\input{figure/8_app_real_robot}

%% file: table/app_robotwin_hyp.tex
\begin{table*}[t]
    \centering
    \caption{Training configurations on RoboTwin 2.0. }
    \label{tab:robotwin_training}
    \scriptsize
    \setlength{\tabcolsep}{3pt}
    \renewcommand{\arraystretch}{1.18}
    \begin{tabular*}{\textwidth}{@{\extracolsep{\fill}}lcccccc@{}}
        \toprule
        Hyperparameter & InterACT & \shortstack{SmolVLA /\\SmolVLA-Duo} & \shortstack{Lila-WAM /\\Lila-WAM-Duo} & \shortstack{FLOWER /\\FLOWER-Duo} & $\pi_{0.5}$-Duo & \shortstack{TwinVLA\\(orig.\ paper)} \\
        \midrule
        Global batch size & 256 & 32 & 1,024 & 512 & 8 & 8 \\
        Optimizer updates & 40,000 & 100,000 & 10,720 & 20,000 & 45,000 & 100,000 \\
        Optimizer & AdamW & AdamW & AdamW & AdamW & AdamW & AdamW \\
        Peak LR (action head) & $10^{-4}$ & $10^{-4}$ & $2\!\times\!10^{-4}$ & $4\!\times\!10^{-4}$ & $2.5\!\times\!10^{-5}$ & $10^{-4}$ \\
        Peak LR (VLM/backbone) & $10^{-5}$ & $10^{-4}$ & Frozen & $8\!\times\!10^{-5}$ & $2.5\!\times\!10^{-5}$ & Frozen \\
        LR schedule & Cosine & Cosine & Cosine & Cosine & Cosine & Cosine \\
        Warmup (updates) & 1,000 & 1,000 & 0 & 500 & 1,000 & 5,000 (5\%) \\
        Minimum LR (\% of peak) & 0\% & 2.5\% & 25\% & 10\% & 10\% & NR \\
        Weight decay & $10^{-4}$ & $10^{-10}$ & 0.01 & 0.01 & 0.01 & $10^{-5}$ \\
        Gradient clipping (max norm) & None & 10 & 1 & 1 & 1 & 1 \\
        Training precision & BF16 AMP & BF16 & BF16 AMP & BF16 AMP & BF16 AMP & BF16 \\
        Action chunk (timesteps) & 50 & 50 & 32 & 50 & 50 & 20 \\
        Random seed & 42 & 1,000 & 0 & 42 & 42 & NR \\
        \bottomrule
    \end{tabular*}
\end{table*}

%% file: table/app_robodojo_hyp.tex
\begin{table*}[t]
    \centering
    \caption{Training configurations on RoboDojo.}
    \label{tab:robodojo_training}
    \scriptsize
    \setlength{\tabcolsep}{3pt}
    \renewcommand{\arraystretch}{1.18}
    \begin{tabular*}{\textwidth}{@{\extracolsep{\fill}}lcccc@{}}
        \toprule
        Hyperparameter & InterACT & SmolVLA & \shortstack{FLOWER /\\FLOWER-Duo} & TwinVLA \\
        \midrule
        Global batch size & 256 & 1,024 & 512 & 128 \\
        Optimizer updates & 30,000 & 36,260 & 20,000 & 48,000 \\
        Optimizer & AdamW & AdamW & AdamW & AdamW \\
        Peak LR (action head) & $10^{-4}$ & $10^{-4}$ & $10^{-4}$ & $2\!\times\!10^{-4}$ \\
        Peak LR (VLM/backbone) & $10^{-5}$ & Frozen & $2\!\times\!10^{-5}$ & Frozen \\
        LR schedule & Cosine & Cosine & Cosine & Cosine \\
        Warmup (updates) & 1,000 & 1,000 & 500 & 250 \\
        Minimum LR (\% of peak) & 0\% & 2.5\% & 0\% & 0\% \\
        Weight decay & $10^{-4}$ & $10^{-10}$ & 0.01 & $10^{-5}$ \\
        Gradient clipping (max norm) & None & 10 & 1 & 1 \\
        Training precision & BF16 AMP & BF16 & BF16 AMP & BF16 \\
        Action chunk (timesteps) & 50 & 50 & 50 & 20 \\
        Random seed & 42 & 0 & 42 & 42 \\
        \bottomrule
    \end{tabular*}
    \par\smallskip
\end{table*}

%% file: table/app_real_hyp.tex
\begin{table*}[t]
    \centering
    \caption{Training configurations for models evaluated on the real-world experiments. }
    \label{tab:real_robot_training}
    \scriptsize
    \setlength{\tabcolsep}{3pt}
    \renewcommand{\arraystretch}{1.18}
    \begin{tabular*}{\textwidth}{@{\extracolsep{\fill}}lcccccc@{}}
        \toprule
        & \multicolumn{2}{c}{Soft-parcel handling} & \multicolumn{2}{c}{Rigid-parcel handling} & \multicolumn{2}{c}{Box packing} \\
        \cmidrule(lr){2-3}\cmidrule(lr){4-5}\cmidrule(l){6-7}
        Hyperparameter & FLOWER-Duo & $\pi_{0.5}$ & FLOWER-Duo & $\pi_{0.5}$ & FLOWER-Duo & $\pi_{0.5}$ \\
        \midrule
        Global batch size & 512 & 512 & 1,024 & 512 & 640 & 256 \\
        Optimizer updates & 8,000 & 8,185 & 8,352 & 8,174 & 19,000 & 10,000 \\
        Optimizer & AdamW & AdamW & AdamW & AdamW & AdamW & AdamW \\
        Peak LR (action head) & $4\!\times\!10^{-4}$ & $2.5\!\times\!10^{-5}$ & $4\!\times\!10^{-4}$ & $2.5\!\times\!10^{-5}$ & $10^{-4}$ & $5\!\times\!10^{-5}$ \\
        Peak LR (VLM/backbone) & $4\!\times\!10^{-5}$ & $2.5\!\times\!10^{-5}$ & $4\!\times\!10^{-5}$ & $2.5\!\times\!10^{-5}$ & $10^{-5}$ & $5\!\times\!10^{-5}$ \\
        LR schedule & Cosine & Cosine & Cosine & Cosine & Cosine & Cosine \\
        Warmup (updates) & 1,000 & 1,000 & 2,000 & 500 & 100 & 500 \\
        Minimum LR (\% of peak) & 10\% & 10\% & 10\% & 10\% & 10\% & 10\% \\
        Weight decay & 0.05 & 0.01 & 0.05 & 0.01 & 0.05 & 0.01 \\
        Gradient clipping (max norm) & 1 & 1 & 1 & 1 & 1 & 1 \\
        Training precision & BF16 AMP & BF16 AMP & BF16 AMP & BF16 AMP & BF16 AMP & BF16 AMP \\
        Action chunk (timesteps) & 50 & 50 & 50 & 50 & 50 & 50 \\
        Random seed & 42 & 42 & 42 & 42 & 42 & 42 \\
        \bottomrule
    \end{tabular*}
    \par\smallskip
\end{table*}

%% file: figure/8_app_real_robot.tex
\begin{figure*}[t]
    \centering
    \includegraphics[width=\textwidth]{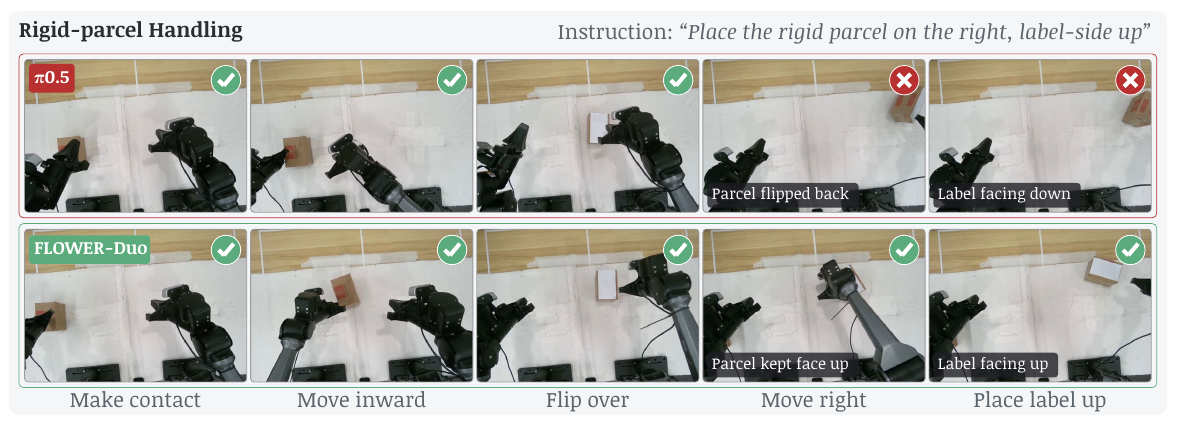}
    \caption{Additional qualitative real-world examples of FLOWER-Duo and $\pi_{0.5}$ on rigid-parcel handling task.}
    \label{fig:app_real_robot}
\end{figure*}